\documentclass[conference]{IEEEtran}
\IEEEoverridecommandlockouts

\usepackage{cite}
\usepackage[utf8]{inputenc}
\usepackage{amsmath}
\usepackage{amssymb}
\usepackage{amsfonts}
\usepackage{algorithmic}
\usepackage{graphicx}
\usepackage{textcomp}
\usepackage{xcolor}
\usepackage[colorlinks=true, linkcolor=blue, citecolor=blue, urlcolor=blue]{hyperref}

\def\BibTeX{{\rm B\kern-.05em{\sc i\kern-.025em b}\kern-.08em
    T\kern-.1667em\lower.7ex\hbox{E}\kern-.125emX}}
\usepackage{romannum}
\usepackage{caption}
\usepackage{subcaption}
\usepackage{tabularx} 
\usepackage{booktabs} 

\begin{document}

\title{Artificial Behavior Intelligence: Technology, Challenges, and Future Directions}

\author{
\IEEEauthorblockN{
Kanghyun Jo\thanks{Corresponding author: Kanghyun Jo (email: acejo@ulsan.ac.kr)}, 
Jehwan Choi, 
Kwanho Kim, 
Seongmin Kim, \\
Duy-Linh Nguyen, 
Xuan-Thuy Vo, 
Adri Priadana, 
Tien-Dat Tran}
\IEEEauthorblockA{
\textit{Department of Electrical, Electronic and Computer Engineering} \\
\textit{University of Ulsan, Ulsan, Korea} \\
acejo@ulsan.ac.kr, cjh1897@ulsan.ac.kr, arrony12@ulsan.ac.kr, dailysmile@ulsan.ac.kr, \\
ndlinh301@mail.ulsan.ac.kr, xthuy@islab.ulsan.ac.kr, priadana@mail.ulsan.ac.kr, ttd9x1995@gmail.com}
}

\maketitle

\begin{abstract}
Understanding and predicting human behavior has emerged as a core capability in various AI application domains such as autonomous driving, smart healthcare, surveillance systems, and social robotics. This paper defines the technical framework of Artificial Behavior Intelligence (ABI), which comprehensively analyzes and interprets human posture, facial expressions, emotions, behavioral sequences, and contextual cues. It details the essential components of ABI, including pose estimation, face and emotion recognition, sequential behavior analysis, and context-aware modeling. Furthermore, we highlight the transformative potential of recent advances in large-scale pretrained models, such as large language models (LLMs), vision foundation models, and multimodal integration models, in significantly improving the accuracy and interpretability of behavior recognition.

Our research team has a strong interest in the ABI domain and is actively conducting research, particularly focusing on the development of intelligent lightweight models capable of efficiently inferring complex human behaviors. This paper identifies several technical challenges that must be addressed to deploy ABI in real-world applications including learning behavioral intelligence from limited data, quantifying uncertainty in complex behavior prediction, and optimizing model structures for low-power, real-time inference. To tackle these challenges, our team is exploring various optimization strategies including lightweight transformers, graph-based recognition architectures, energy-aware loss functions, and multimodal knowledge distillation, while validating their applicability in real-time environments.

\end{abstract}

\section{Introduction}
The philosopher Aristotle once described human beings as “social animals.” This statement implies that humans do not exist as isolated entities, but rather live in constant interaction and communication with others. Humans intuitively perceive others’ emotions, states, and intentions through their tone of voice, facial expressions, gestures, and behavioral patterns. These abilities are fundamental to mutual understanding and empathetic social interaction. In other words, humans can interpret the underlying meaning of others’ behaviors and predict their subsequent actions. This cognitive ability is now referred to as Behavior Intelligence.
Behavior Intelligence is an essential cognitive function for understanding and responding to the complexities of human interaction. It involves inferring another person’s state, desires, and motivations to generate contextually appropriate responses. This capacity enables smooth communication, cooperation, and coexistence in human society.
With the rapid advancement of artificial intelligence (AI), machines have evolved beyond basic information processing and are now capable of generating and interpreting diverse data such as language, images, and video. This progress has made it possible to develop intelligent interactive agents that move beyond scripted responses and can produce linguistically and non-linguistically appropriate reactions based on context and situation. Within this paradigm shift, we propose the need for Artificial Behavior Intelligence a system that understands and anticipates human behavior.
In this paper, we define Artificial Behavior Intelligence as “the computer-implemented cognitive ability to understand the geometric and abstract meanings embedded in human behavior by considering national, cultural, and situational contexts, and to infer likely future behaviors.” This goes beyond mere action classification or detection; it refers to higher-order intelligence capable of grasping the background and intent behind behaviors and predicting subsequent actions. For example, an autonomous vehicle can infer the intent of a pedestrian to cross the street based on their gesture, or a service robot can recognize a user's emotional state from facial expressions and body language and respond accordingly.
There are numerous reasons why Behavior Intelligence is essential. In domains such as autonomous driving, smart healthcare, surveillance systems, social robotics, and virtual assistants, the ability to understand and respond to human behavior significantly enhances both safety and user experience. Moreover, Behavior Intelligence holds the potential to address pressing societal issues such as elderly care in aging populations, mental health monitoring, and crowd management in urban environments.
In this context, the structure of this paper is as follows. Chapter 2 compares and analyzes the global definitions and trends related to Behavior Intelligence. Chapter 3 examines the key technologies needed to realize Behavior Intelligence, such as pose estimation, emotion recognition, and sequential behavior analysis. Chapter 4 explores the impact of emerging generative AI technologies including LLMs and large vision models (LVMs) on ABI, and proposes future research directions. Finally, Chapter 5 provides a comprehensive summary of the current status and future prospects of Behavior Intelligence, discussing its broader social and technological implications.

\begin{table*}[!ht]
\centering
\caption{Definition and Application Areas of Behavioral Intelligence}
\label{tab:behavioral_intelligence}
\begin{tabularx}{\textwidth}{>{\bfseries}l X X}
\toprule
Perspective & Definition and Description of Behavioral Intelligence & Application Areas \\
\midrule
Academia (Behavior Informatics) &
An interdisciplinary field that models, represents, analyzes, and manages individual or group behaviors to derive behavioral intelligence and insights. Aims to comprehensively understand behavioral decision-making processes without self-report bias. \cite{fourth_kluai} &
Behavioral analysis, healthcare, marketing strategy \\
\addlinespace
Industry (Behavior AI) &
Refers to AI systems that learn patterns from past behavioral data to analyze and predict human behavior. Emphasizes real-time adaptation to complex and dynamic interactions, with a focus on behavior prediction under changing circumstances. \cite{fifth_dragonspears_behavioral_ai} &
Autonomous driving, robotics, surveillance systems, sports analytics \\
\addlinespace
Psychology (Behavioral Intelligence) &
The capacity to perceive and interpret one's own and others’ behavior across various situations to understand and influence social interactions. This intelligence interprets and predicts behavior based on internal and external cues, playing a critical role in modern interpersonal and occupational contexts. &
Interpersonal relationships, emotional regulation, empathy \\
\bottomrule
\end{tabularx}
\end{table*}

\section{Behavior Intelligence}

\subsection{Diverse Definitions and Conceptual Spectrum of Behavior Intelligence}
The term Behavior Intelligence has not yet been standardized and is interpreted in various ways depending on the field and the researcher. Disciplines such as psychology, computer vision, sociology, and artificial intelligence each apply the concept within their unique contexts, reflecting both the complexity of the term and its wide applicability.
For example, in psychology, behavior intelligence is often regarded as “the ability to recognize and appropriately respond to one’s own behavior and the behavior of others in diverse social situations” \cite{first_Vernon01061990}. It is typically classified as a subset of social intelligence, complementing traditional concepts such as IQ (intelligence quotient) and EQ (emotional quotient), and is closely related to an individual's capacity for social adaptation. Furthermore, it is described as an essential skill in interpersonal relationships and professional settings in modern society \cite{second_Taillard2013}.
In contrast, in the fields of computer science and artificial intelligence, behavior intelligence refers to a machine’s ability to observe, understand, and further predict or respond to human behaviors. Initially focused on vision-based action recognition, this concept has evolved toward understanding abstract meanings and social contexts underlying behaviors. It includes the capacity to interpret continuous behaviors along a temporal axis and infer the intent or emotional state behind specific actions. Notably, Fuchs et al. (2023) \cite{third_10.1145/3580492} define behavior intelligence as “the ability to flexibly adapt and guide actions in human-machine interactions depending on context,” and propose that realizing such intelligence requires an integrated approach involving game theory, reinforcement learning, and Theory of Mind (ToM).
As shown, behavior intelligence is approached differently across disciplines. Table 1 summarizes the characteristics of behavior intelligence as defined in each domain, along with their primary application areas.

In particular, Behavioral AI has recently emerged as a prominent concept in the industrial sector. It predicts future behaviors and supports automated decision-making based on various forms of digital behavioral data, such as clicks, browsing patterns, movement trajectories, and gaze tracking. This represents a notable example in which behavior intelligence has evolved beyond being a mere analytical tool into an applied technology that directly contributes to tangible value creation.

\begin{figure*}[!ht]
    \centering
    \includegraphics[width=0.9\linewidth]{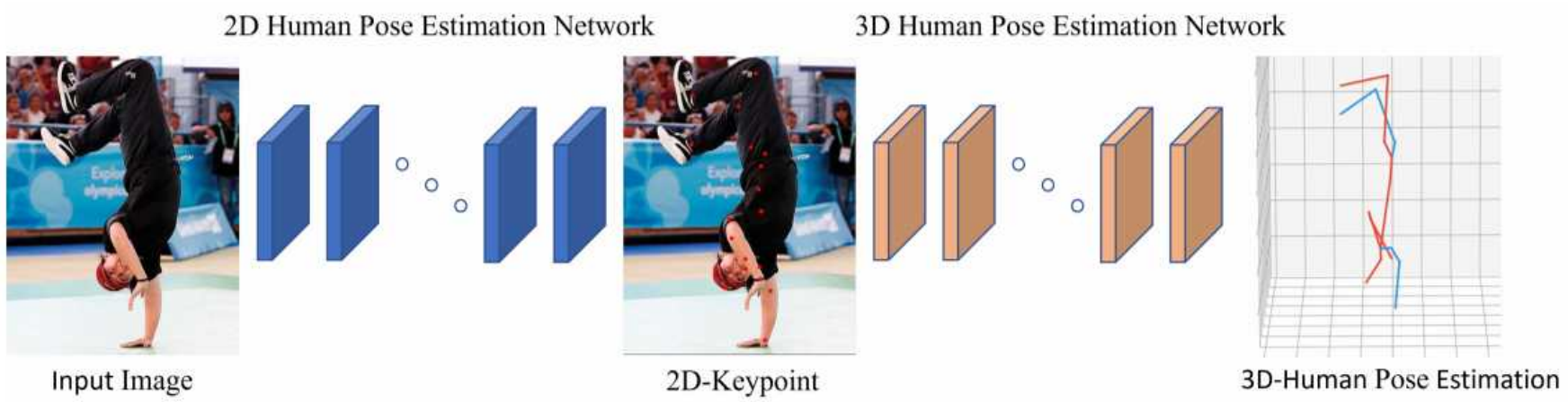}
    \caption{Inference Process of 2D and 3D Pose Estimation \cite{thirteenth_10003954}}
    \label{fig:taehwa_bridge_image}
\end{figure*}
\begin{figure*}[!ht]
    \centering
    \includegraphics[width=0.9\linewidth]{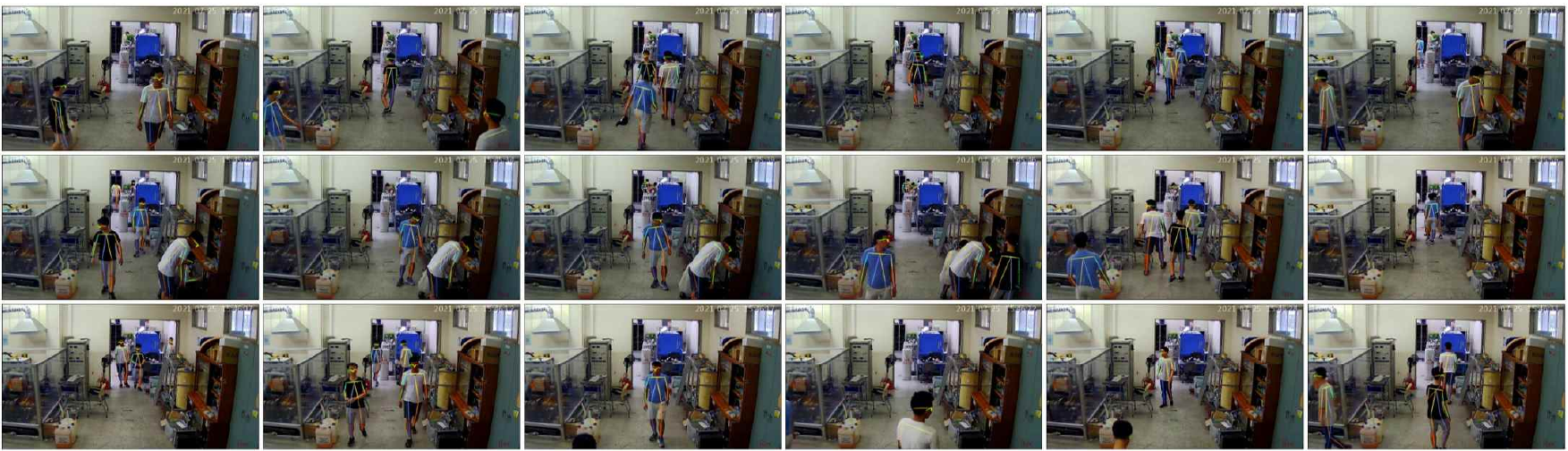}
    \caption{Inference Result Pose Estimation \cite{thirteenth_10003954}}
    \label{fig:taehwa_bridge_image}
\end{figure*}

\subsection{Variations in Perspectives Across Nations and Cultures}
The definition and application of behavior intelligence can vary depending on national and cultural contexts. This is because the same behavior may carry different meanings across cultural regions. For example, in many Asian countries, it is common to wave a hand or call out loudly to get a waiter’s attention. However, in some Western countries, such behavior may be perceived as rude or inappropriate. Likewise, nodding one's head generally signals agreement in Western cultures, but in certain Asian countries, it can simply indicate that the listener is paying attention, regardless of agreement.
Therefore, to implement behavior intelligence effectively in global environments, it is essential to design systems that recognize and incorporate cross-cultural differences. In this context, recent research has emphasized the need for Culturally Cognizant AI—artificial intelligence capable of understanding cultural nuances \cite{sixth_prabhakaran2022culturalincongruenciesartificialintelligence} This is especially crucial for interactive agents such as robots or chatbots that must interact with users around the world, as they must learn and adapt to different behavioral norms and expectations depending on the region.
Individual countries have also begun to explore the concept of behavior intelligence from their own national perspectives. A notable example is the Electronics and Telecommunications Research Institute (ETRI) of South Korea, which has developed a technology that allows devices to autonomously determine necessary actions and carry out tasks, i.e. a technology they call ActionBrain. When applied to robotics, this system enables machines to observe and imitate the behaviors of human workers, learn through virtual simulations, and ultimately generate behavior intelligence to support collaborative manufacturing tasks. \cite{seventh_nst_actionbrain} This represents a practical implementation of behavior intelligence, where robots in industrial settings understand human behavior and respond with appropriate actions.
In North America, Professor Alex Pentland at the MIT Media Lab has conducted extensive research on understanding human behavior using network science. A representative project involves the Sociometer \cite{seventh_nst_actionbrain}, a wearable sensor used to measure face-to-face human interactions. His work analyzes the structure and dynamics of human communication networks, offering a method to extract patterns and intentions embedded in human behavior from data. Such efforts form part of the broader field of behavioral intelligence research, uncovering actionable behavioral insights for diverse real-world applications.

\subsection{Deriving the Core Components of Behavior Intelligence}
Synthesizing the discussions in Sections 2.1 and 2.2, it becomes evident that contextual understanding, behavioral pattern learning, and future behavior prediction are fundamental components of behavior intelligence. These core elements can be articulated individually as follows:
Contextual Understanding:
Behavior intelligence goes beyond simple action recognition to interpret the meaning of behavior by considering national, cultural, and situational contexts. Since the same behavior may have different meanings across cultures, AI systems must recognize and adapt to such differences. Previous research emphasizes that when AI systems mimic human behavior, they must take into account the embedded cultural context: Failure to do so may lead to misunderstandings or errors in culturally diverse environments. Therefore, culturally aware AI is essential for the realization of behavior intelligence.
Behavioral Pattern Analysis:
AI systems must identify and model patterns from historical behavioral data to predict future actions. As the ability to collect and analyze large-scale behavioral data increases, AI systems gain deeper insights into human behavior that were previously unattainable. The concept of Internet of Behaviors (IoB) \cite{eighth_vt_iob_ai} exemplifies this, where vast streams of behavioral data gathered from IoT devices and sensors are analyzed by AI to understand and even influence the “when, how, and why” of human actions.
Prediction and Application:
The ultimate goal of behavior intelligence is to predict near-future behavior and adapt in real-time to dynamic contexts. For example, AI-based behavior prediction models can assess a driver’s acceleration and braking patterns to evaluate safe driving behavior, which may then be used to offer discounted insurance premiums \cite{eighth_vt_iob_ai}. Similarly, online streaming services have advanced to the point where they can infer a user’s current mood and recommend content accordingly. These examples show that behavior intelligence aims not only for statistical prediction but for intelligent inference that holistically considers both individual states and surrounding environments.

\section{Detailed Technology}
The implementation of behavior intelligence requires several core technological components, including pose estimation, facial recognition, emotion recognition, sequential behavior analysis, and context modeling. In recent years, each of these fields has seen significant advancement through the introduction of various deep learning–based techniques. This section introduces the key technologies, current research trends, and representative methodologies essential for enabling behavior intelligence.

\subsection{Pose Estimation}
Pose estimation is a foundational technique for understanding human behavior, involving the detection of body joint locations from images or videos. Since the introduction of deep learning, 2D pose estimation methods based on Convolutional Neural Networks (CNNs) have significantly outperformed traditional approaches. For example, multi-scale CNN architectures such as the Hourglass Network and OpenPose, as well as high-resolution representation models like HRNet, have demonstrated high accuracy in detecting human joint keypoints \cite{ninth_Lan_2023}.
More recently, Vision Transformer–based models utilizing self-attention mechanisms have been proposed. These models capture global spatial relationships within a scene, further improving pose estimation performance in video settings \cite{tenth_967c68400625467f827c45f8129d5a35, thirteenth_10003954}. Research on 3D pose estimation is also gaining momentum, with models capable of predicting three-dimensional joint coordinates from a single RGB camera using deep neural networks, as well as techniques that fuse multi-view camera data to accurately reconstruct 3D human poses \cite{eleventh_NOGUEIRA2025105437}.
In parallel, Graph Neural Network (GNN)–based approaches have emerged to learn structural relationships among joints. Notably, skeletal graphs represent joints as nodes, and Graph Convolutional Networks (GCNs) are used to refine pose estimation results or recognize actions from pose sequences \cite{fourteenth_10377200}.
Alongside performance improvements, research on lightweight and real-time pose estimation is gaining attention. Some studies have introduced efficient architectures that incorporate lightweight spatial-attention modules, achieving both reduced computational cost and increased accuracy \cite{twelfth_10595678}. As a result, recent lightweight models have enabled real-time inference even on CPUs or mobile devices. However, challenges remain in accurately estimating poses in complex scenes involving multiple individuals or in handling occlusion scenarios.

\begin{figure}[!h]
    \centering
    \includegraphics[width=1\linewidth]{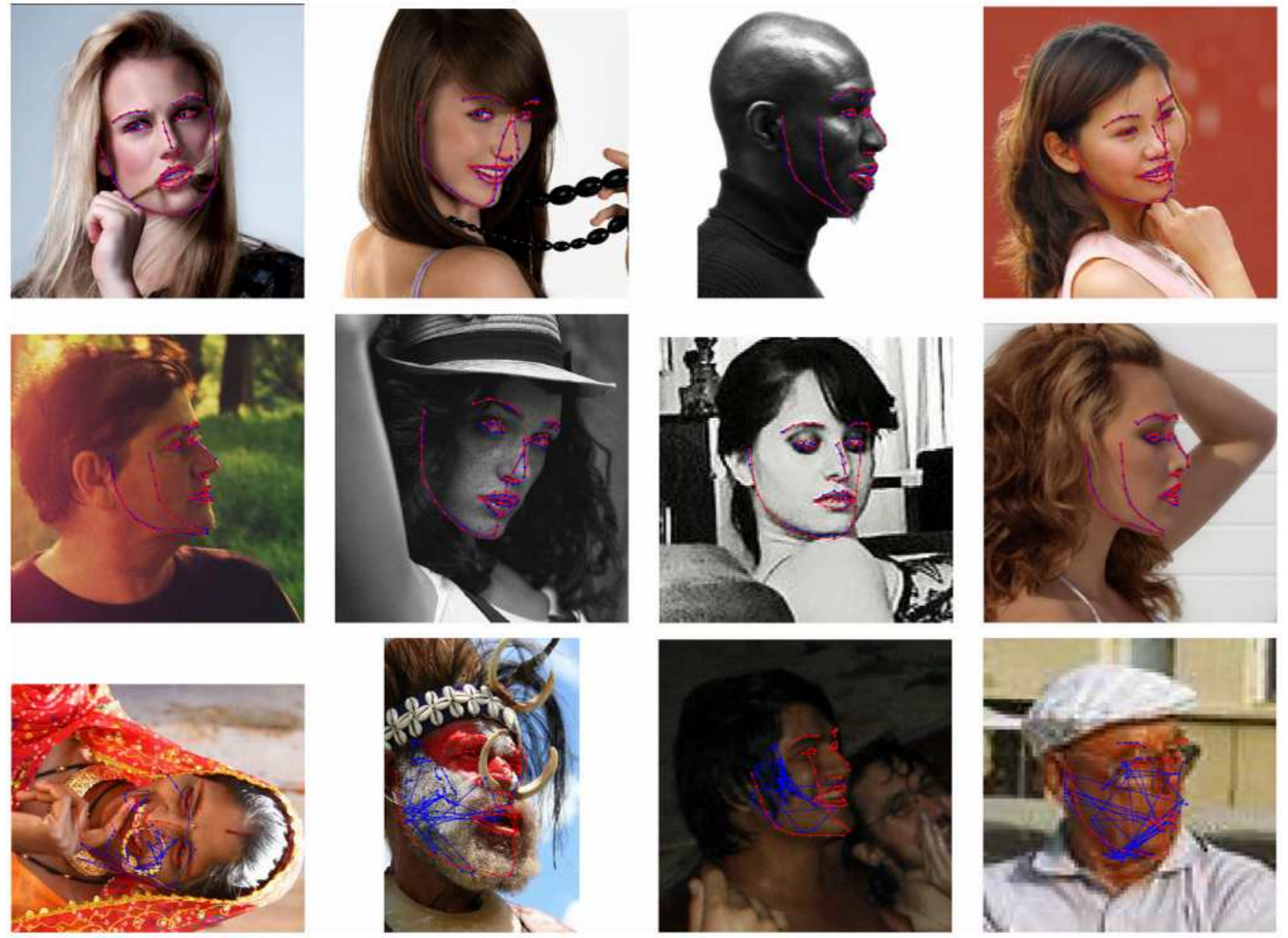}
    \caption{The result of face recognition \cite{eighteenth_8959132}}
    \label{fig:taehwa_bridge_image}
\end{figure}

\begin{figure}[!h]
    \centering
    \includegraphics[width=1\linewidth]{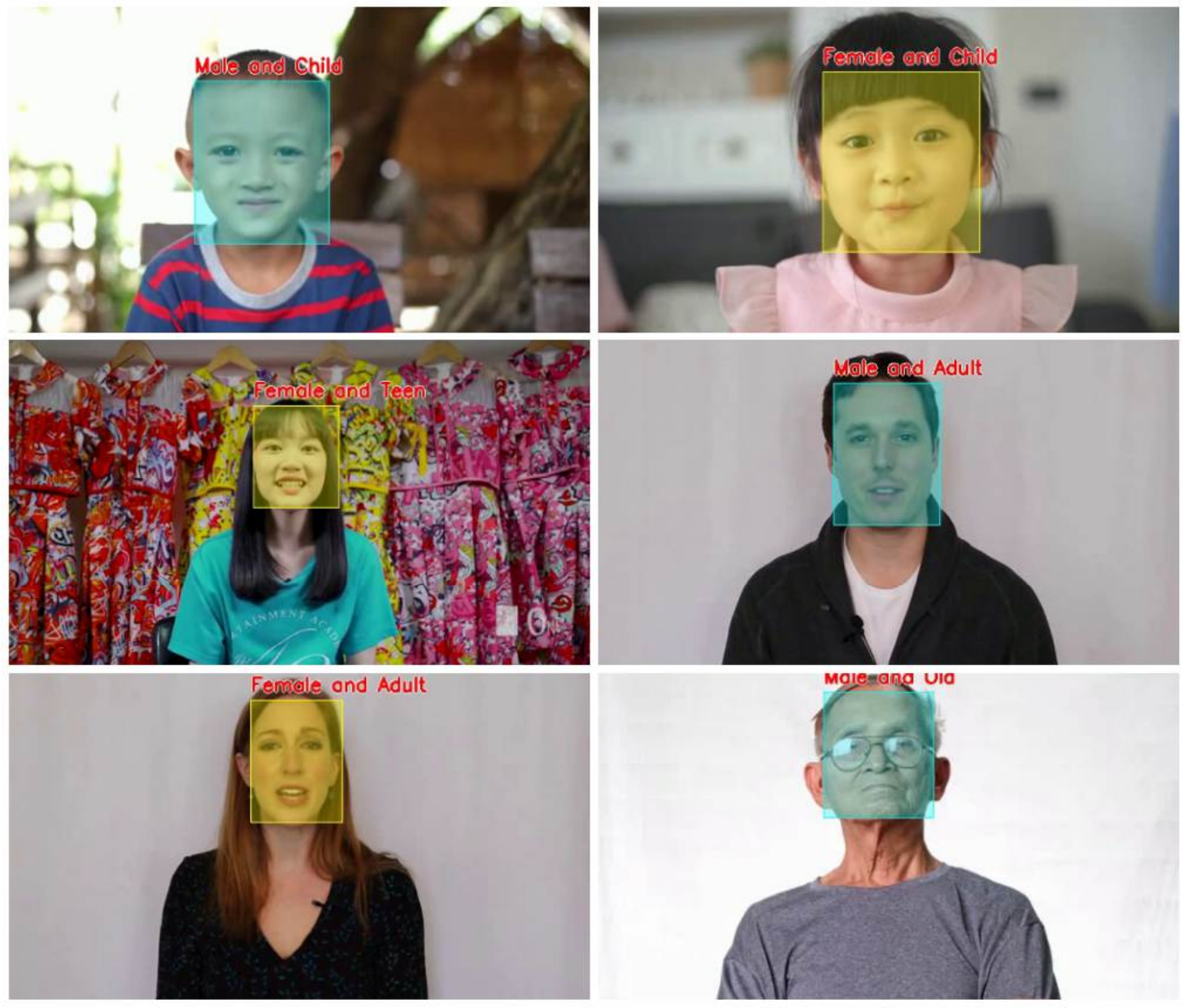}
    \caption{The result of age estimation \cite{nineteenth_10595810}}
    \label{fig:taehwa_bridge_image}
\end{figure}

\subsection{Face Recognition}
Face recognition is a technology used to detect or identify individuals by analyzing facial features from input images or video. It is widely applied in domains such as security authentication, video surveillance, and Human-Robot Interaction (HRI). Deep learning–based face recognition typically involves training deep CNN embedding models on large-scale facial datasets, and then comparing the distances between embeddings to determine identity—achieving performance comparable to that of humans.
Notably, models such as ArcFace \cite{fifteenth_8953658} have achieved over 99\% accuracy on benchmarks like Labeled Faces in the Wild (LFW) by improving both the loss function and network architecture. Subsequent models, including CosFace \cite{sixteenth_8578650} and SphereFace \cite{seventeenth_8100196}, have introduced stable and effective methods for learning discriminative face embeddings.
More recently, research has explored Vision Transformer–based face recognition using self-attention mechanisms, as well as integrated face tracking across multi-camera and video streams. Few-shot and zero-shot face recognition, which aim to recognize individuals from limited or no prior samples, have also gained traction, often through meta-learning or pretrained model approaches. For example, some studies apply the multimodal pretrained model CLIP to perform zero-shot face classification tasks.
\begin{figure*}[!ht]
    \centering
    \includegraphics[width=1\linewidth]{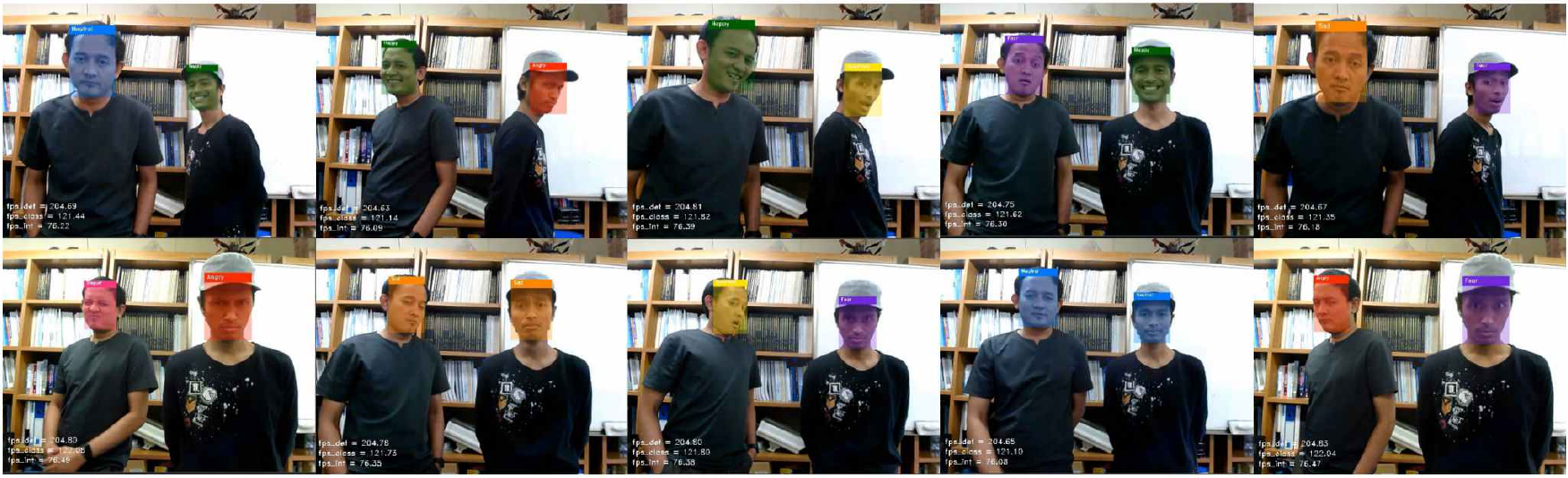}
    \caption{Emotion Recognition Result using Edge Device(CPU) \cite{23_10753436}}
    \label{fig:taehwa_bridge_image}
\end{figure*}
Accurate face recognition often requires prior face detection and alignment (i.e., facial landmark detection). Recent advances in 3D facial landmark estimation have enabled more robust normalization across pose variations \cite{eighteenth_8959132}. Additional research has utilized lightweight Transformer encoders for tasks such as facial attribute prediction and age estimation \cite{nineteenth_10595810, twentieth_10.1007/978-981-97-4249-3_9}, contributing to broader facial analysis capabilities. These techniques, when combined with face recognition pipelines, further enhance overall system performance.
Despite strong performance in identifying individuals under controlled conditions, face recognition systems still face challenges under varying conditions such as mask-wearing, low lighting, or low-resolution video. To address these issues, recent studies have proposed methods such as data augmentation using Generative Adversarial Networks (GANs), domain adaptation techniques, and 3D face modeling to extract pose-invariant features.

\subsection{Emotion Recognition}
Emotion recognition refers to the technology that infers a person’s emotional state using cues such as facial expressions, voice, and gestures. Traditionally, facial expression analysis has been the core focus. However, with the rise of multimodal emotion recognition, recent approaches integrate facial video with other modalities such as speech, body movements, and physiological signals to infer emotional states more comprehensively.
Early deep learning–based facial expression recognition models used deep CNNs like VGG and ResNet to classify expressions. More recent models incorporate attention mechanisms to emphasize emotionally salient facial regions. One representative example is Dual Attention Face Emotion Recognition (FER) \cite{22_9254805}. Another notable system is EMOTIZER \cite{23_10753436}, a multi-pose emotion recognition system capable of robust performance even from various facial angles. EMOTIZER combines a lightweight CNN for face detection and a dedicated network for expression classification, using both global and multi-response attention modules to achieve high accuracy in complex backgrounds. Impressively, it operates at 76 FPS on a CPU, demonstrating significant progress in model efficiency.
Beyond facial expressions, growing research explores emotion recognition from body language and vocal cues. For example, some approaches analyze pose sequences to detect emotional states from body posture changes, while others interpret vocal features such as intonation and speech rate to infer emotions. This shift reflects a broader trend toward leveraging the full spectrum of multimodal emotional signals \cite{21_https://doi.org/10.1002/widm.1563}.
Recent deep learning studies highlight the importance of effective feature alignment and fusion across these modalities. Key research areas include temporal synchronization between modalities, shared representation learning, suppression of irrelevant features, and emphasis on discriminative features. Despite high accuracy in controlled lab environments, emotion recognition systems often suffer from performance degradation in real-world settings due to factors like lighting variation, occlusion, and individual differences.
To address these limitations, ongoing research focuses on collecting large-scale affective datasets, developing domain generalization techniques, and training models that incorporate cultural variations in emotional expression.

\subsection{Sequential Human Behavior Analysis}
Sequential human behavior analysis refers to the interpretation or prediction of behavior patterns over time. This includes video-based action recognition, behavior forecasting, and anomaly detection. Traditionally, video-based behavior recognition has utilized models such as LSTM or RNN to integrate frame-level 2D CNN features along the temporal axis, or 3D convolutional networks (Conv3D) to learn spatiotemporal features simultaneously.
Recently, video transformer models incorporating the Transformer architecture have been introduced for sequential behavior analysis. For instance, TimeSformer \cite{24_bertasius2021spacetimeattentionneedvideo} and ViViT \cite{25_arnab2021vivit} employ self-attention mechanisms to capture both temporally significant frames and spatially relevant regions, achieving state-of-the-art performance. In the domain of skeleton-based action recognition, Graph Convolutional Networks (GCNs) have become a standard approach. Starting with ST-GCN \cite{26_Yan_Xiong_Lin_2018}, which classifies pose sequences using spatiotemporal graph convolutions, numerous extended studies have been proposed that improve graph structures or incorporate attention mechanisms. A notable example is HD-GCN \cite{27_Lee_2023_ICCV}, which utilizes the hierarchical structure of human skeletal graphs to achieve state-of-the-art accuracy in skeleton-based behavior recognition. The integration of GCNs and attention mechanisms \cite{28_twentyeighth_chen2025} has significantly advanced the accurate understanding of continuous actions.
\begin{figure}[!ht]
    \centering
    \includegraphics[width=1\linewidth]{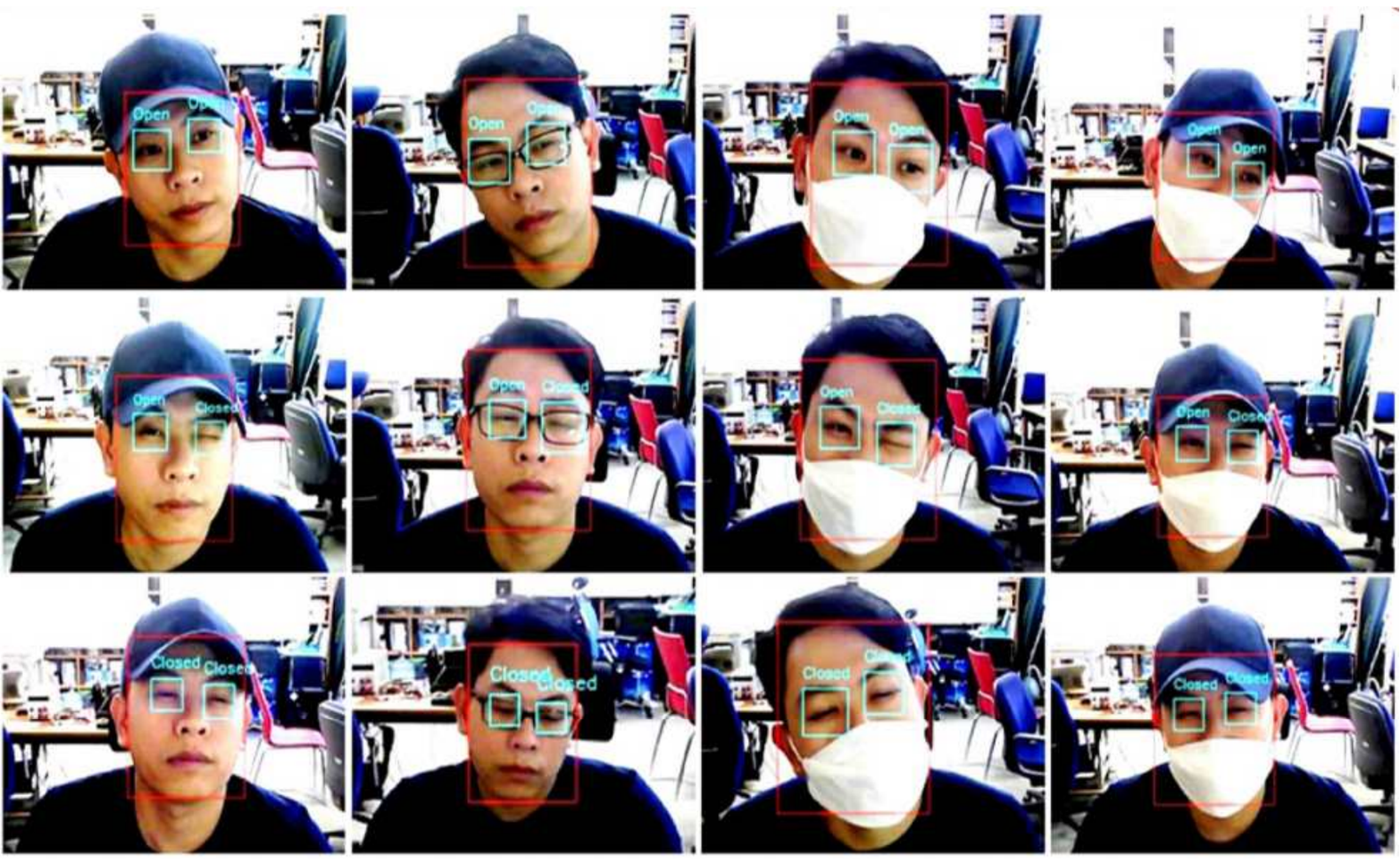}
    \caption{Driver Eye Status Surveillance System using Sequential Human Behavior Analysis \cite{29_10223702}}
    \label{fig:taehwa_bridge_image}
\end{figure}
There is also growing interest in zero-shot and few-shot behavior recognition, which aims to identify novel behaviors that are not included in the training dataset. This research direction is motivated by the challenge of labeling complex and diverse action videos at scale. ActionCLIP, for example, applies the large-scale image-text model CLIP to video data, enabling zero-shot action classification through textual descriptions without requiring additional training. With increasing model complexity and representation power in deep learning, sequential action analysis has achieved a level of expressiveness far beyond earlier capabilities.
Such advances have led to the development of lightweight driver behavior classification models that combine convolutional networks and attention mechanisms to detect inattentive actions like phone use, smoking, or drowsiness while driving \cite{29_10223702, 30_9810252}. These studies demonstrate the feasibility of real-time behavior analysis systems and have evolved into AI models capable of detecting the start and end points of specific actions in video streams, or predicting the next action based on observed behavioral sequences.

\subsection{Context and Object Recognition}
Context modeling refers to the technology that interprets human behavior in relation to its surrounding environment, enabling higher-level reasoning in behavior intelligence. Since the meaning of the same action can vary depending on context, AI systems must incorporate contextual information—such as environmental cues, interaction targets, and prior activities—in order to accurately interpret human behavior.
Recent studies have employed graph-based reasoning and attention-based context integration techniques to model the relationships between people and objects in a scene or to understand the context of social interactions. For example, in Human-Object Interaction (HOI) recognition, models construct graphs where humans and objects serve as nodes, and the edges represent learned interactions (e.g., holding, lying down) \cite{31_9489275}.
When multiple people are present in a scene, social graphs and contextual attention mechanisms are used to interpret factors such as relative distances, gaze directions, and gestures—allowing for a deeper understanding of individual actions within a crowd. In \cite{32_9816965}, a real-time human detection model for robotic assistance was proposed that utilizes multi-stage contextual blocks. This method combines background information with hierarchical contextual cues to quickly detect even small objects by leveraging environmental features (e.g., floor-wall intersections), thereby improving detection confidence.
Moreover, context modeling can be extended to incorporate temporal context. This involves analyzing how previous actions relate causally or sequentially to the current behavior. To model this, approaches such as memory networks and temporal attention mechanisms are being studied, where representations of past actions are stored and recalled to enhance current behavior recognition in continuous action analysis.

\section{Challenges}
The implementation and real-world application of behavior intelligence technologies face several technical and theoretical challenges. The key issues can be summarized as follows: the difficulty of data collection and labeling, performance discrepancies across cultural domains, limitations in emotion and intention recognition, uncertainty in prediction outcomes, constraints on real-time processing, and ethical concerns.

\subsection{Data Collection and Labeling}
Training behavior intelligence models requires vast amounts of behavioral data, yet capturing and labeling complex human behaviors is an extremely time- and cost-intensive process \cite{33_TONG2025126147}. Although a number of public datasets have been developed, such as UCF101 and Kinetics for human actions, and FER2013 for facial emotions, efforts to gather more data through self-recording and crowd-sourcing have been continuing.
To reduce the labeling burden, approaches such as self-supervised learning are being employed to pretrain models on video features, and synthetic data generation (e.g., creating virtual scenarios using game engines) is also being tracted. However, challenges remain in comprehensively collecting complex behavioral data, leading to increased interest in few-shot and weakly supervised learning techniques \cite{34_wanyan2025comprehensivereviewfewshotaction}.
Despite these advancements, building large-scale datasets that capture the full diversity of real-world behavior patterns remains an unresolved task. Moreover, issues such as dataset bias continue to hinder the generalizability and fairness of trained models.

\subsection{Accuracy in Emotion and Intention Recognition}
Recognizing internal emotions and intentions—beyond surface-level behaviors—is a difficult task, even for humans. Although technologies for estimating emotions through facial expressions or actions have significantly advanced, they still struggle to accurately interpret subtle emotional states (e.g., ambiguous smiles, mixed emotions) \cite{35_BISOGNI2023104724}, and exhibit limitations in reading intentions that emerge through interactions with others.
Deep learning–based emotion recognition has achieved over 90\% accuracy on well-curated datasets \cite{36_thirtysixth_elsheikh2024}, but this performance typically applies to clear and distinct expressions. In real world environments, accuracy can drop to around 60\% \cite{37_thirtyseventh_agung2024}. Intention recognition, which combines video processing and prediction models, remains particularly challenging, as intention is a latent variable that cannot be directly observed, making it difficult to measure or trust the results.
Research efforts have explored indirect intention inference through gaze tracking, gesture interpretation, and contextual simulation, but studies have concluded that far more progress is needed before machines can reach human-level understanding \cite{33_TONG2025126147}. As a result, while coarse classification of emotional or intentional states is now feasible, the fine-grained recognition of nuanced emotions or complex decision-making processes remains an open challenge.

\subsection{Prediction Uncertainty}
Although the outputs of deep learning models are expressed as probability values, it remains to evaluate how those predictions are reliable for unresolved issues. Therefore, it is necessary to devise such a method assessing uncertainty in behavior recognition results and enabling models to detect when they are less confident.
Some recent studies have explored uncertainty estimation using Bayesian deep learning and ensemble techniques. For example, methods such as Monte Carlo Dropout have been proposed to approximate output distributions, and the variance across multiple model predictions has been used to assess uncertainty \cite{38_4f2e753556e94addb686c9f787ac6393}. However, most behavior recognition models still rely on single-point predictions, making it difficult to respond effectively when errors occur—especially when the input data distribution differs significantly from the training data or contains noise. In such cases, deep learning models tend to fail to obtain certainty. 
For instance, medical image analysis has begun incorporating model uncertainty estimates into the output for human decision \cite{40_SALVI2025105846}. Similarly, such approaches need to be adopted in future human behavior recognition systems. There is no standardized method for quantifying uncertainty in deep learning, which presents a major challenge in ensuring the safety and reliability of behavior intelligence systems.

\subsection{Constraints in Real-Time Processing}
It is necessary to analyze and estimate human behavior in various situations, including healthcare, sports, and security. In these cases, accuracy and real-time responsiveness are critical. To recognize and respond to human behavior in real-time, it is essential to have an algorithmic speed optimization.
With advancements in hardware performance and the widespread use of GPU acceleration, the inference speed of deep learning models for behavior analysis has improved significantly compared to the past \cite{ninth_Lan_2023}. For example, \cite{23_10753436} demonstrated that with lightweight models, tasks such as simple pose estimation or facial expression recognition for one or two individuals in a single video stream can be processed in real-time, achieving 76 FPS on a CPU.
To further optimize performance, techniques such as knowledge distillation, network pruning/quantization, and Neural Architecture Search (NAS) have been adopted to find efficient model structures. These methods have successfully reduced inference latency in some applications.
Nonetheless, challenges remain in scenarios involving high-resolution video streams or simultaneous recognition of multiple subjects. For instance, recognizing behaviors while tracking many individuals in a crowd requires a substantial increase in computational load, often leading to frame rate degradation unless high-performance GPUs are used. In the case of Multi-Person Pose Estimation (MPPE), computational complexity scales with the number of individuals, meaning that even lightweight models designed for single-person inference may fail to maintain real-time performance in multi-person settings.
Moreover, embedded and mobile environments impose power consumption constraints, making it difficult to deploy standard deep learning models directly. While ongoing research in model compression and the development of Edge AI accelerator chips has mitigated some of these limitations, achieving fully real-time understanding of complex behaviors without latency remains an unresolved challenge.

\section{Future Technology}
The future of behavior intelligence is expected to be significantly advanced through the integration of foundation models, which are rapidly emerging as transformative technologies. In particular, the advent of Large Language Models (LLMs), Large Vision Models (LVMs), and multimodal integrated models opens new possibilities for deeper understanding of human behavior and contextual awareness.
The incorporation of LLMs enables AI to perform higher-level semantic analysis and reasoning about human behavior. Since LLMs possess extensive embedded knowledge and commonsense reasoning capabilities, they can linguistically interpret behaviors captured in video and infer underlying context. For instance, while a conventional behavior recognition model might simply label someone repeatedly checking their watch as “looking at a watch,” an LLM could infer and describe this as “anxiously waiting for someone,” considering the surrounding context. Such explainable behavior reasoning, powered by advanced natural language processing and knowledge-based inference, holds promise for simulating social interactions.
In \cite{41_stanford_generative_agents}, a system known as Generative Agents was introduced based on an LLM like ChatGPT. In this framework, virtual agents form memories, make plans, and interact with one another in a way that mimics daily human life. This system demonstrated that LLMs can go beyond recognizing human behavior to generating and simulating realistic social behavior \cite{42_10.1145/3586183.3606763}. Accordingly, future behavior intelligence systems are expected to leverage large models to understand intentions, generate appropriate response strategies, and engage in human-like interactive conversations.
Large Vision Models (LVMs) and multimodal foundation models also offer broad visual understanding capabilities critical for human behavior recognition. For example, OpenAI’s CLIP model \cite{43_radford2021learningtransferablevisualmodels}, trained on massive image-text pairs from the internet, has shown strong conceptual understanding of visual content. When applied to behavior videos, CLIP enables zero-shot action recognition \cite{44_QUAN2025111402}, meaning the model can interpret behaviors without having seen labeled examples, by relying on textual descriptions. These vision-language models can help alleviate data scarcity and support more flexible and adaptive behavior recognition.
In the vision domain, general-purpose models like the Vision Transformer (ViT) and Segment Anything Model (SAM) are receiving increasing attention, while in video analysis, research on large-scale Video Foundation Models is progressing \cite{45_wang2024internvideo2scalingfoundationmodels}. Google’s PaLM-E \cite{46_10.5555/3618408.3618748}, introduced in 2023, is a notable example of a multimodal model combining a large language model with robotic vision and sensor inputs. This model demonstrated that a single architecture could process multiple sensory modalities and generate robotic control commands \cite{33_TONG2025126147}, suggesting that large multimodal models can be used in Embodied AI to span everything from scene understanding to behavioral decision-making.
Ultimately, future behavior intelligence systems will integrate visual, auditory, and linguistic inputs through large multimodal models to achieve fine-grained situational understanding and behavior analysis. As part of this high-level behavior comprehension, research on emotion inference and future behavior generation is also essential. As discussed, LLMs and multimodal models may be capable not only of recognizing surface-level behavior but also of inferring psychological states. In near future, we may see emotionally intelligent AI assistants that synthesize cues such as facial expressions, language, and behavior to infer a person’s current emotions and intentions—a form of machine empathy. This would enable the realization of emotionally responsive agents that understand and adapt to user feelings.
Furthermore, in terms of predicting and generating future behaviors, models will evolve to generate plausible behavior scenarios based on past behavioral data. While such generative functions were once partially implemented through probabilistic models or simulators, foundation models are expected to directly generate and recommend human-like behavior patterns. In \cite{42_10.1145/3586183.3606763}, for example, virtual agents in a simulated town spontaneously created new social events, spread information, and formed relationships, closely mirroring real human interactions. In the future, AI may reach the level of simulating emergent social behaviors, capable of imitating and predicting complex human interactions.
In summary, behavior intelligence will evolve beyond simple behavior identification into a holistic intelligence that understands context, empathizes with emotion, predicts the future, and responds or interacts appropriately.

\section{Conclusion}
This paper has explored Artificial Behavior Intelligence, a computational implementation of the human ability to understand social interactions and interpret behaviors. Humans possess an innate capacity to intuitively perceive emotions, intentions, and contextual cues from others' actions. Recent advances in artificial intelligence have shown the potential for machines to replicate these high-level cognitive capabilities. In particular, behavior intelligence is becoming a crucial technological foundation across diverse domains such as autonomous vehicles, smart healthcare, robotics, surveillance systems, and virtual assistants.
To realize behavior intelligence, we identified core enabling technologies: Pose Estimation, Face Recognition, Emotion Recognition, Sequential Human Behavior Analysis, and Contextual Modeling. Each of these fields has experienced significant breakthroughs by development in deep learning, Transformer architectures, Graph Neural Networks (GNNs), and multimodal fusion techniques. Nonetheless, the challenges remain in the difficulty of data collection and labeling, the limited accuracy of emotion and intention recognition, prediction uncertainty, and real-time processing constraints. In addition, mathematical limitations reside in complex optimization problems for 3D pose estimation, the sensitivity of face recognition to lighting, angles, and resolution, and the subjectivity and variability of emotional expressions.
Despite these limitations, the emergence of Large Language Models (LLMs), Large Vision Models (LVMs), and integrated multimodal models presents a powerful opportunity for advancement in human behavior intelligence. These foundation models enable new capabilities such as explainable reasoning, zero-shot recognition, simulation of social interactions, and higher-level understanding and prediction of emotions and intentions. As a result, behavior intelligence is expected to evolve beyond simple behavior classification, transforming into a comprehensive and human-centered AI technology that can understand psychological states, empathize, predict future behaviors, and respond naturally.
For continued progress in behavior intelligence, future research must address issues such as data bias, ethical implications, and technological uncertainty. It is still necessary to overcome these challenges for development of AI systems, improving quality of life(QoL), contributing to a safer and more convenient society.



\begin{thebibliography}{10}
\providecommand{\url}[1]{#1}
\csname url@samestyle\endcsname
\providecommand{\newblock}{\relax}
\providecommand{\bibinfo}[2]{#2}
\providecommand{\BIBentrySTDinterwordspacing}{\spaceskip=0pt\relax}
\providecommand{\BIBentryALTinterwordstretchfactor}{4}
\providecommand{\BIBentryALTinterwordspacing}{\spaceskip=\fontdimen2\font plus
\BIBentryALTinterwordstretchfactor\fontdimen3\font minus \fontdimen4\font\relax}
\providecommand{\BIBforeignlanguage}[2]{{%
\expandafter\ifx\csname l@#1\endcsname\relax
\typeout{** WARNING: IEEEtran.bst: No hyphenation pattern has been}%
\typeout{** loaded for the language `#1'. Using the pattern for}%
\typeout{** the default language instead.}%
\else
\language=\csname l@#1\endcsname
\fi
#2}}
\providecommand{\BIBdecl}{\relax}
\BIBdecl

\bibitem{fourth_kluai}
``Behavior informatics,'' \url{https://klu.ai/glossary/behavior-informatics}, accessed: 2025-05-06.

\bibitem{fifth_dragonspears_behavioral_ai}
S.~Provvidenza, ``The future of behavioral ai: Predictions and emerging trends,'' \url{https://www.dragonspears.com/blog/future-of-behavioral-ai}, 2024, accessed: 2025-05-06.

\bibitem{first_Vernon01061990}
\BIBentryALTinterwordspacing
P.~A.~V. and, ``The use of biological measures to estimate behavioral intelligence,'' \emph{Educational Psychologist}, vol.~25, no. 3-4, pp. 293--304, 1990. [Online]. Available: \url{https://doi.org/10.1080/00461520.1990.9653115}
\BIBentrySTDinterwordspacing

\bibitem{second_Taillard2013}
\BIBentryALTinterwordspacing
M.~Taillard and H.~Giscoppa, \emph{Behavioral Intelligence}.\hskip 1em plus 0.5em minus 0.4em\relax New York: Palgrave Macmillan US, 2013, pp. 183--187. [Online]. Available: \url{https://doi.org/10.1057/9781137347329_19}
\BIBentrySTDinterwordspacing

\bibitem{third_10.1145/3580492}
\BIBentryALTinterwordspacing
A.~Fuchs, A.~Passarella, and M.~Conti, ``Modeling, replicating, and predicting human behavior: A survey,'' \emph{ACM Trans. Auton. Adapt. Syst.}, vol.~18, no.~2, May 2023. [Online]. Available: \url{https://doi.org/10.1145/3580492}
\BIBentrySTDinterwordspacing

\bibitem{thirteenth_10003954}
T.-D. Tran, X.-T. Vo, D.-L. Nguyen, and K.-H. Jo, ``Combination of deep learner network and transformer for 3d human pose estimation,'' in \emph{2022 22nd International Conference on Control, Automation and Systems (ICCAS)}, 2022, pp. 174--178.

\bibitem{sixth_prabhakaran2022culturalincongruenciesartificialintelligence}
\BIBentryALTinterwordspacing
V.~Prabhakaran, R.~Qadri, and B.~Hutchinson, ``Cultural incongruencies in artificial intelligence,'' 2022. [Online]. Available: \url{https://arxiv.org/abs/2211.13069}
\BIBentrySTDinterwordspacing

\bibitem{seventh_nst_actionbrain}
Electronics and Telecommunications Research Institute (ETRI), ``ETRI Develops 'Action Brain' That Enables Objects to Act on Their Own,'' ((in Korean) \url{https://www.nst.re.kr/www/selectBbsNttView.do?key=62&bbsNo=13&nttNo=14971}, 2024. Accessed: 2025-05-06.


\bibitem{eighth_vt_iob_ai}
{Virginia Tech}, ``Ai and the internet of behaviors: Exploring the intersection of data, technology, and human behavior,'' \url{https://vtmit.vt.edu/academics/student-experience/blog/ai-internet-behaviors.html}, 2023, accessed: 2025-05-06.

\bibitem{ninth_Lan_2023}
\BIBentryALTinterwordspacing
G.~Lan, Y.~Wu, F.~Hu, and Q.~Hao, ``Vision-based human pose estimation via deep learning: A survey,'' \emph{IEEE Transactions on Human-Machine Systems}, vol.~53, no.~1, p. 253–268, Feb. 2023. [Online]. Available: \url{http://dx.doi.org/10.1109/THMS.2022.3219242}
\BIBentrySTDinterwordspacing

\bibitem{tenth_967c68400625467f827c45f8129d5a35}
A.~Ulhaq, N.~Akhtar, G.~Pogrebna, and A.~Mian, ``\BIBforeignlanguage{English}{Vision transformers for action recognition: A survey},'' arXiv, United States, WorkingPaper, Sep. 2022, 15 Figures and 5 Tables.

\bibitem{eleventh_NOGUEIRA2025105437}
\BIBentryALTinterwordspacing
A.~F.~R. Nogueira, H.~P. Oliveira, and L.~F. Teixeira, ``Markerless multi-view 3d human pose estimation: A survey,'' \emph{Image and Vision Computing}, vol. 155, p. 105437, 2025. [Online]. Available: \url{https://www.sciencedirect.com/science/article/pii/S0262885625000253}
\BIBentrySTDinterwordspacing

\bibitem{fourteenth_10377200}
J.~Lee, M.~Lee, D.~Lee, and S.~Lee, ``Hierarchically decomposed graph convolutional networks for skeleton-based action recognition,'' in \emph{2023 IEEE/CVF International Conference on Computer Vision (ICCV)}, 2023, pp. 10\,410--10\,419.

\bibitem{twelfth_10595678}
T.-D. Tran, G.~Cao, R.~Ashraf, and K.-H. Jo, ``Group spatial attention for 3d human pose estimation,'' in \emph{2024 IEEE 33rd International Symposium on Industrial Electronics (ISIE)}, 2024, pp. 1--7.

\bibitem{eighteenth_8959132}
V.-T. Hoang, D.-S. Huang, and K.-H. Jo, ``3-d facial landmarks detection for intelligent video systems,'' \emph{IEEE Transactions on Industrial Informatics}, vol.~17, no.~1, pp. 578--586, 2021.

\bibitem{nineteenth_10595810}
A.~Priadana, D.-L. Nguyen, X.-T. Vo, M.~D. Putro, G.~Cao, and K.~Jo, ``Simultaneous facial age group and gender recognition using efficient local-global attention network for intelligent advertising,'' in \emph{2024 IEEE 33rd International Symposium on Industrial Electronics (ISIE)}, 2024, pp. 1--6.

\bibitem{fifteenth_8953658}
J.~Deng, J.~Guo, N.~Xue, and S.~Zafeiriou, ``Arcface: Additive angular margin loss for deep face recognition,'' in \emph{2019 IEEE/CVF Conference on Computer Vision and Pattern Recognition (CVPR)}, 2019, pp. 4685--4694.

\bibitem{sixteenth_8578650}
H.~Wang, Y.~Wang, Z.~Zhou, X.~Ji, D.~Gong, J.~Zhou, Z.~Li, and W.~Liu, ``Cosface: Large margin cosine loss for deep face recognition,'' in \emph{2018 IEEE/CVF Conference on Computer Vision and Pattern Recognition}, 2018, pp. 5265--5274.

\bibitem{seventeenth_8100196}
W.~Liu, Y.~Wen, Z.~Yu, M.~Li, B.~Raj, and L.~Song, ``Sphereface: Deep hypersphere embedding for face recognition,'' in \emph{2017 IEEE Conference on Computer Vision and Pattern Recognition (CVPR)}, 2017, pp. 6738--6746.

\bibitem{23_10753436}
M.~Dwisnanto~Putro, A.~Priadana, D.-L. Nguyen, and K.-H. Jo, ``Emotizer: A multipose facial emotion recognizer using rgb camera sensor on low-cost devices,'' \emph{IEEE Sensors Journal}, vol.~25, no.~2, pp. 3708--3718, 2025.

\bibitem{twentieth_10.1007/978-981-97-4249-3_9}
A.~Priadana, D.-L. Nguyen, X.-T. Vo, and K.~Jo, ``Human facial age group recognizer using assisted bottleneck transformer encoder,'' in \emph{Frontiers of Computer Vision}, G.~Irie, C.~Shin, T.~Shibata, and K.~Nakamura, Eds.\hskip 1em plus 0.5em minus 0.4em\relax Singapore: Springer Nature Singapore, 2024, pp. 108--121.

\bibitem{22_9254805}
M.~D. Putro, D.-L. Nguyen, and K.-H. Jo, ``A dual attention module for real-time facial expression recognition,'' in \emph{IECON 2020 The 46th Annual Conference of the IEEE Industrial Electronics Society}, 2020, pp. 411--416.

\bibitem{21_https://doi.org/10.1002/widm.1563}
\BIBentryALTinterwordspacing
M.~P.~A. Ramaswamy and S.~Palaniswamy, ``Multimodal emotion recognition: A comprehensive review, trends, and challenges,'' \emph{WIREs Data Mining and Knowledge Discovery}, vol.~14, no.~6, p. e1563, 2024. [Online]. Available: \url{https://wires.onlinelibrary.wiley.com/doi/abs/10.1002/widm.1563}
\BIBentrySTDinterwordspacing

\bibitem{24_bertasius2021spacetimeattentionneedvideo}
\BIBentryALTinterwordspacing
G.~Bertasius, H.~Wang, and L.~Torresani, ``Is space-time attention all you need for video understanding?'' 2021. [Online]. Available: \url{https://arxiv.org/abs/2102.05095}
\BIBentrySTDinterwordspacing

\bibitem{25_arnab2021vivit}
A.~Arnab, M.~Dehghani, G.~Heigold, C.~Sun, M.~Lu{\v{c}}i{\'c}, and C.~Schmid, ``Vivit: A video vision transformer,'' in \emph{International Conference on Computer Vision (ICCV)}, 2021.

\bibitem{26_Yan_Xiong_Lin_2018}
\BIBentryALTinterwordspacing
S.~Yan, Y.~Xiong, and D.~Lin, ``Spatial temporal graph convolutional networks for skeleton-based action recognition,'' \emph{Proceedings of the AAAI Conference on Artificial Intelligence}, vol.~32, no.~1, Apr. 2018. [Online]. Available: \url{https://ojs.aaai.org/index.php/AAAI/article/view/12328}
\BIBentrySTDinterwordspacing

\bibitem{27_Lee_2023_ICCV}
J.~Lee, M.~Lee, D.~Lee, and S.~Lee, ``Hierarchically decomposed graph convolutional networks for skeleton-based action recognition,'' in \emph{Proceedings of the IEEE/CVF International Conference on Computer Vision (ICCV)}, October 2023, pp. 10\,444--10\,453.

\bibitem{28_twentyeighth_chen2025}
\BIBentryALTinterwordspacing
D.~Chen, M.~Chen, P.~Wu \emph{et~al.}, ``Two-stream spatio-temporal gcn-transformer networks for skeleton-based action recognition,'' \emph{Scientific Reports}, vol.~15, p. 4982, 2025. [Online]. Available: \url{https://doi.org/10.1038/s41598-025-87752-8}
\BIBentrySTDinterwordspacing

\bibitem{29_10223702}
D.-L. Nguyen, M.~D. Putro, and K.-H. Jo, ``Lightweight cnn-based driver eye status surveillance for smart vehicles,'' \emph{IEEE Transactions on Industrial Informatics}, vol.~20, no.~3, pp. 3154--3162, 2024.

\bibitem{30_9810252}
------, ``Driver behaviors recognizer based on light-weight convolutional neural network architecture and attention mechanism,'' \emph{IEEE Access}, vol.~10, pp. 71\,019--71\,029, 2022.

\bibitem{31_9489275}
H.~Wang, L.~Jiao, F.~Liu, L.~Li, X.~Liu, D.~Ji, and W.~Gan, ``Ipgn: Interactiveness proposal graph network for human-object interaction detection,'' \emph{IEEE Transactions on Image Processing}, vol.~30, pp. 6583--6593, 2021.

\bibitem{32_9816965}
M.~D. Putro, D.-L. Nguyen, A.~Priadana, and K.-H. Jo, ``Fast person detector with efficient multi-level contextual block for supporting assistive robot,'' in \emph{2022 IEEE 5th International Conference on Industrial Cyber-Physical Systems (ICPS)}, 2022, pp. 1--6.

\bibitem{33_TONG2025126147}
\BIBentryALTinterwordspacing
T.~Tong, R.~Setchi, and Y.~Hicks, ``Human intention recognition using context relationships in complex scenes,'' \emph{Expert Systems with Applications}, vol. 266, p. 126147, 2025. [Online]. Available: \url{https://www.sciencedirect.com/science/article/pii/S0957417424030148}
\BIBentrySTDinterwordspacing

\bibitem{34_wanyan2025comprehensivereviewfewshotaction}
\BIBentryALTinterwordspacing
Y.~Wanyan, X.~Yang, W.~Dong, and C.~Xu, ``A comprehensive review of few-shot action recognition,'' 2025. [Online]. Available: \url{https://arxiv.org/abs/2407.14744}
\BIBentrySTDinterwordspacing

\bibitem{35_BISOGNI2023104724}
\BIBentryALTinterwordspacing
C.~Bisogni, L.~Cimmino, M.~{De Marsico}, F.~Hao, and F.~Narducci, ``Emotion recognition at a distance: The robustness of machine learning based on hand-crafted facial features vs deep learning models,'' \emph{Image and Vision Computing}, vol. 136, p. 104724, 2023. [Online]. Available: \url{https://www.sciencedirect.com/science/article/pii/S0262885623000987}
\BIBentrySTDinterwordspacing

\bibitem{36_thirtysixth_elsheikh2024}
\BIBentryALTinterwordspacing
R.~A. Elsheikh, M.~A. Mohamed, A.~M. Abou-Taleb \emph{et~al.}, ``Improved facial emotion recognition model based on a novel deep convolutional structure,'' \emph{Scientific Reports}, vol.~14, p. 29050, 2024. [Online]. Available: \url{https://doi.org/10.1038/s41598-024-79167-8}
\BIBentrySTDinterwordspacing

\bibitem{37_thirtyseventh_agung2024}
\BIBentryALTinterwordspacing
E.~S. Agung, A.~P. Rifai, and T.~Wijayanto, ``Image-based facial emotion recognition using convolutional neural network on emognition dataset,'' \emph{Scientific Reports}, vol.~14, p. 14429, 2024. [Online]. Available: \url{https://doi.org/10.1038/s41598-024-65276-x}
\BIBentrySTDinterwordspacing

\bibitem{38_4f2e753556e94addb686c9f787ac6393}
\BIBentryALTinterwordspacing
F.~Verdoja and V.~Kyrki, ``\BIBforeignlanguage{English}{Notes on the behavior of mc dropout},'' Jul. 2021, iCML Workshop on Uncertainty \&amp; Robustness in Deep Learning, ICML UDL ; Conference date: 23-07-2021 Through 23-07-2021. [Online]. Available: \url{https://sites.google.com/view/udlworkshop2021}
\BIBentrySTDinterwordspacing

\bibitem{40_SALVI2025105846}
\BIBentryALTinterwordspacing
M.~Salvi, S.~Seoni, A.~Campagner, A.~Gertych, U.~Acharya, F.~Molinari, and F.~Cabitza, ``Explainability and uncertainty: Two sides of the same coin for enhancing the interpretability of deep learning models in healthcare,'' \emph{International Journal of Medical Informatics}, vol. 197, p. 105846, 2025. [Online]. Available: \url{https://www.sciencedirect.com/science/article/pii/S1386505625000632}
\BIBentrySTDinterwordspacing

\bibitem{41_stanford_generative_agents}
\BIBentryALTinterwordspacing
{Stanford Human-Centered Artificial Intelligence}, ``Computational agents exhibit believable humanlike behavior,'' September 2023, accessed: 2025-05-06. [Online]. Available: \url{https://hai.stanford.edu/news/computational-agents-exhibit-believable-humanlike-behavior}
\BIBentrySTDinterwordspacing

\bibitem{42_10.1145/3586183.3606763}
\BIBentryALTinterwordspacing
J.~S. Park, J.~O'Brien, C.~J. Cai, M.~R. Morris, P.~Liang, and M.~S. Bernstein, ``Generative agents: Interactive simulacra of human behavior,'' in \emph{Proceedings of the 36th Annual ACM Symposium on User Interface Software and Technology}, ser. UIST '23.\hskip 1em plus 0.5em minus 0.4em\relax New York, NY, USA: Association for Computing Machinery, 2023. [Online]. Available: \url{https://doi.org/10.1145/3586183.3606763}
\BIBentrySTDinterwordspacing

\bibitem{43_radford2021learningtransferablevisualmodels}
\BIBentryALTinterwordspacing
A.~Radford, J.~W. Kim, C.~Hallacy, A.~Ramesh, G.~Goh, S.~Agarwal, G.~Sastry, A.~Askell, P.~Mishkin, J.~Clark, G.~Krueger, and I.~Sutskever, ``Learning transferable visual models from natural language supervision,'' 2021. [Online]. Available: \url{https://arxiv.org/abs/2103.00020}
\BIBentrySTDinterwordspacing

\bibitem{44_QUAN2025111402}
\BIBentryALTinterwordspacing
Z.~Quan, J.~Chen, D.~Deguchi, J.~Sun, C.~Zhang, Y.~Li, and H.~Murase, ``Semantic matters: A constrained approach for zero-shot video action recognition,'' \emph{Pattern Recognition}, vol. 162, p. 111402, 2025. [Online]. Available: \url{https://www.sciencedirect.com/science/article/pii/S0031320325000627}
\BIBentrySTDinterwordspacing

\bibitem{45_wang2024internvideo2scalingfoundationmodels}
\BIBentryALTinterwordspacing
Y.~Wang, K.~Li, X.~Li, J.~Yu, Y.~He, C.~Wang, G.~Chen, B.~Pei, Z.~Yan, R.~Zheng, J.~Xu, Z.~Wang, Y.~Shi, T.~Jiang, S.~Li, H.~Zhang, Y.~Huang, Y.~Qiao, Y.~Wang, and L.~Wang, ``Internvideo2: Scaling foundation models for multimodal video understanding,'' 2024. [Online]. Available: \url{https://arxiv.org/abs/2403.15377}
\BIBentrySTDinterwordspacing

\bibitem{46_10.5555/3618408.3618748}
D.~Driess, F.~Xia, M.~S.~M. Sajjadi, C.~Lynch, A.~Chowdhery, B.~Ichter, A.~Wahid, J.~Tompson, Q.~Vuong, T.~Yu, W.~Huang, Y.~Chebotar, P.~Sermanet, D.~Duckworth, S.~Levine, V.~Vanhoucke, K.~Hausman, M.~Toussaint, K.~Greff, A.~Zeng, I.~Mordatch, and P.~Florence, ``Palm-e: an embodied multimodal language model,'' in \emph{Proceedings of the 40th International Conference on Machine Learning}, ser. ICML'23.\hskip 1em plus 0.5em minus 0.4em\relax JMLR.org, 2023.

\end{thebibliography}
\end{document}